Bayesian Inference in Model-Based Machine Vision

Thomas O. Binford*, Tod S. Levitt**, and Wallace B. Mann*

* Robotics Laboratory, Stanford University
** Advanced Decision Systems

Abstract

We present a thorough integration of hierarchical Bayesian inference with comprehensive physical representation of objects and their relations in a system for reasoning with geometry in machine vision. Bayesian inference provides a framework for accruing probabilities to rank order hypotheses. This is a preliminary version of visual interpretation in SUCCESSOR, an intelligent, model-based vision system integrating multiple sensors.

## Introduction

Our design for machine vision uses an evidential accrual process, a. beginning from representation and database of a priori models of physical objects and their photometric, geometric, and functional properties, together with their relationships and environment, b. predicting observables using models of sensors and perceptual measurement processes; c. making measurements of corresponding observables, measuring image evidence for features of objects and structures of features such as edges, vertices and regions; d. generating hypotheses of instances of objects from those measurements and predictions; and f. accumulating evidence in support or denial of those hypotheses, matching that evidence against the models to arrive at interpretations of the imaged world.

Physical objects are 3d. In range imagery, measurements are 3d. There is still a difficult stage of segmenting and estimating 3d relations that disclose object structure. In 2d images, there is an additional inference from 2d projected image evidence to 3d interpretation of surfaces. System structure tends to break up into a natural hierarchy of representation and processing [Binford 80].

In hierarchical vision system representation, objects are recursively broken up into parts. This part-of hierarchy representation fits naturally into a Bayesian inference framework [Levitt 86]. A hierarchical representation for Bayesian inference consists of alternating layers of parts and joints, relationships between parts. Such a hierarchy forms a directed acyclic graph (DAG), where nodes are parts or relations. Direction between nodes indicates "causation" in the sense that a parent-part "causes" its sub-parts and sub-part relations to exist. A DAG for machine vision based on generalized cylinders (GCs, described below) as the primitive 3D volume representation is pictured in figures 1a and 1b. Figure 1a shows the DAG accruing evidence up to GCs, while figure 1b shows the DAG from GCs up through the top level objects modeled in the vision system.

Inference for machine vision is an ambiguous and errorful process because the evidence provided in an image does not map one to one into the space of possible object models. Evidence in support or denial of a given object is always partial and sometimes incorrect due to noise and/or compounding of errorful interpretation algorithms. On the other hand, there is typically an abundance of evidence [Lowe 84]. One approach to handling this situation is to introduce numerical methods for accruing belief in support or denial of hypotheses of physical objects and their relationships. Then conflicting interpretations can be rank ordered. Bayesian inference provides a framework that is guaranteed to be mathematically coherent, and thus provides a sound basis for system modeling.

Figure 2 shows the structure of our model-based vision system. SUCCESSOR uses generic 3d symbolic object models in contrast to image models (phenomenological models). It includes a powerful modeling subsystem [Ponce 87a]. SUCCESSOR has a prediction module to make symbolic predictions of appearances of objects. There is a generic observability model for imagery in the visible spectrum, along with models of sensors and perceptual operators.

Key issues that motivate our design are:
a. generality: Interpretation capability should be insensitive to variations within object class including articulation and amputation; insensitive to variations in observation, i.e. in viewpoint, sensor type, illumination, and environment, including obscuration; and the system should handle generic, partial interpretation in terms of class models, as well as recognition of instances of highly constrained objects.
b. statistical efficiency: Interpretation models should be as accurate as available perceptual evidence supports, and no more, i.e. including accurate statistical models of uncertainty.
c. computational efficiency: Computational cost should be limited by hypothesis management, especially hypothesis generation.

Related prior work in these directions includes: hypothesis generation and structure matching of [Nevatia 74]; Bayesian perceptual strategies of [Garvey 76]; constraint satisfaction of [Brooks 81], quasi-invariant visual observables in [Binford 79], and hierarchical inference in [Levitt 86]. A discussion of intelligent vision systems and design concepts for general vision systems appeared in [Binford 82]. The experience of using ACRONYM in the Intelligent Task Automation (ITA) project has affected SUCCESSOR design [Chelberg 84].

## Generic Interpretation

Interpretation objectives for the system can be classified along several axes: 1. generic or individual objects; 2. generic or single observation, i.e. with viewpoint-insensitive or viewpoint-sensitive methods; 3. many or few objects or classes. By far the majority of systems choose individual, viewpoint-sensitive, and few objects for the three choices. Our aim for SUCCESSOR is generic objects, generic observations, and many objects.

An example is recognizing a one-armed, bald man as a human being, despite the wide variation among humans or among different observations of the same human, variations in clothing, hairstyle, beard, skin color, growth, age and sex. The issue is similarity, rather than identity in matching. Thus, we seek to match instances of imaged objects at coarse, i.e., more generic, levels of the taxonomic is-a hierarchy, as well as at the level of individual models. A central theme of our approach is geometric representation as a basis for abstractions of shape defining similarity. Parameterized structured models, as in parameters of GCs and parameters of the relations between them, allow matching data to weaker object classes.

Our objective is to be capable of interpreting observed objects using a very large visual memory of models of objects. [Nevatia 74] demonstrated efficient hypothesis generation, selecting subclasses of similar objects from a structured visual memory by shape indexing using coarse, stick-figure, structural descriptions of figures. Recognition strategies in most current vision systems use brute force matching, comparing each model with each observed object, a method that is plausible only with a few objects.

## Geometric Modeling

Object models are physical models; their geometry is represented by part/whole graphs of generalized cylinder primitives. The subset of generalized cylinders implemented is not uniform throughout the system. For display, closed form analytic solutions exist for limbs (apparent edges) of star-shaped, straight, homogeneous GCs (SHGC) including solids of revolution, twisted GCs, and tubes [Ponce 87a]. Generalized cylinders (GCs) are defined by a cross section swept along a space curve, the axis, under a sweeping transformation. The cross section need not be planar, the axis need not be straight or planar, and the sweep may include non-rigid deformations. Figure 3a shows a true cylinder, a cross section swept constant along a straight line. Figure 3b shows a straight homogeneous GC; it has a straight axis and sweeping rule with variable scaling along the axis. Figure 3c is the cross section for 3a and 3b. Figure 3d is the sweeping rule for 3b. Twisted GCs are generated by a cross section that is twisted as it is swept along the axis. Pipes are generated by a constant circular cross section swept along a space curve. By incorporating deformations, a very large class of GCs can be displayed. However, set operations combining these primitives are now computed only for straight homogeneous GCs, SHGCs, that are considered for the remainder of the paper.

Set intersection is computed by a hierarchical, iterative refinement, adaptive algorithm [Ponce 87b]. A tree of boxes encloses each surface; if the surfaces intersect, boxes in the trees intersect. Where the boxes intersect, they are subdivided to refine intersection. The intersection computation cost is $O(\sqrt{n})$.

Primitives are built with an interactive spline editor to specify the cross sections and sweeping rules of our primitives as cubic splines, with knots that are $C^0$, $C^1$, or $C^2$. Compound object models are graphs of primitives represented in a simple modelling language. Affixment arcs are labelled by a transformation, a rotation is represented by Euler angles and a translation represented by its three coordinates $x$, $y$, $z$ in the reference frame of the part. Transformations between objects may be symbolic expressions, allowing articulation. Relative positions of nodes may be defined by spatial relations with expressions like "with face $< face1 >$ in contact with $< face2 >$". The affixment and spatial relation graph are trees, so the relative positions of objects are uniquely determined. We aim to utilize observations of surface finish, paint, and surface texture. Material modeling is based on optical properties, reflectivities, specularities and color [Healey 87a,b].

## Prediction

Figure 4 illustrates the prediction process. Prediction takes the graph of parts and produces a graph of surfaces and their intersections, edges. Intersections, holes, and affixments between surfaces are inferred to determine hidden surfaces and boundaries. Work in progress includes modeling surface material and finish for predictions of color and specularity. Predictions are structured according to object structure, to support perceptual strategies. If viewpoint is completely known, surface predictions are exact and hidden surfaces are suppressed. If viewpoint is unknown, all surfaces are in the graph. Predictions include symbolic expressions with variables for viewpoint, etc. Predictions are made for ribbons, i.e. projections of generalized cylinders, for vertices, and for edges and limbs. A major emphasis is on prediction and hypothesis generation based on quasi-invariants, which we motivate in the following.

For convex polyhedra, each surface is visible over half the viewing sphere. Over most of this hemisphere, quasi-invariants for that plane are nearly constant. Predicting surfaces is $O(n)$ in the number of surfaces $n$. It is straight-forward to include articulations, obscurations, and parameter variations within object class.

However, if we try to predict total views, there are many of them. A paradigm is to predict a dense set of images to use for matching. Dense set is defined by the choice of features to match and the matching algorithm. In SUCCESSOR, prediction from object models differs strongly from this paradigm and similar, such as aspect graphs. The aspect graph approach is to subdivide the viewing sphere into sectors from which the projected topology is invariant. For polyhedra, each time we cross a face there is a new view. Consequently, there are close to $2^n$ views for $n$ faces. It is prohibitive in the aspect graph approach to include articulation, obscuration, and object parameter variation. For general vision, the aspect graph is not a computationally attractive approach for us. Instead, we use quasi-invariant observations as the basis for prediction and matching.

Quasi-invariants provide a theory for computing the probabilistic support that observations lend to inference of object/relationship hypotheses. In the following, we first describe the structure of our Bayesian inference, and then present the theory of quasi-invariants.

## Bayesian Inference

We use probabilities in machine vision to: a. accrue numerical belief for the match of runtime evidence to a priori models of physical objects and relations; b. rank order conflicting interpretations; c. and to feed back this quantified ranking as an input to control and focus of attention in vision system processing.

To apply Bayesian inference to accrue the probability of an interpretation of imaged objects, we must decide, a priori, which modeled objects and relationships we could confuse on the basis of evidence extracted from observables. This is equivalent to specifying the criteria by which object/relationship hypotheses will be generated at runtime. We then form a DAG in which nodes represent all possible conflicting hypotheses based on a common set of evidence sources and possible parent models.

We use a standard Bayes-net formulation to accrue probabilities over this network. However, we vary from traditional implementations in that we do not instantiate nodes in the network until evidence is available to support their existence. This is necessary in machine vision [Levitt 86] because of the inherent combinatorics of the visual domain. Also, a forest of non-conflicting DAGs may be generated at runtime to interpret spatially separated clusters of visual evidence. We do not directly address control issues in this paper.

We briefly summarize the Bayes-net methodology. For a more thorough presentation, see [Pearl 86]. At each node of the DAG in Figure 1, there is a probability distribution over the set of mutually exclusive and exhaustive possible interpretations of the visual evidence accrued to that level in the hierarchy. Each alternate interpretation, e.g. GC, t-joint, valve, is a hypothesis about the environmental situation. Thus a node is a set of hypotheses, e.g. t-joint versus elbow-joint verus non-joint. Although they do not have to be simultaneously instantiated, the possible links between nodes are hard-wired, a priori, by the models of objects and relationships, and the criteria for node instantiation that determines which pieces of evidence can generate conflicting hypotheses. Each alternative hypothesis at a node contributes some probability to the truth of an alternative hypothesis at a parent node (e.g. the part supports the existence of the whole) and also contributes to the truth of supporting children. When new evidence appears at a node, it is assimilated and appropriate versions of that evidence are propagated along all other links entering or exiting the node. Propagation algorithms include necessary computations to avoid double-counting evidence throughout the net. Evidence propagates in time proportional to the diameter of the instantiated net (see [Pearl 86]).

One of the main reasons to use a hierarchical approach like the DAG of figure 1 is to break down the evidential accrual process both to correspond to the process of matching between evidence and models, and to make it feasible to compute prior probabilities between nodes in the DAG. A basic assumption in this approach is that given a parent node, A, with child links, B and C, the probabilities of alternatives at B and C are independent when conditioned on the existence of an alternative in the parent node A. That is, $p(B(i), C(j)|A(k)) = p(B(i)|A(k)) * P(C(j)|A(k))$, where the subscripts indicate alternative hypotheses at the nodes. Conditional independence under this condition is a faithful model of reality for machine vision, because the child-parent relationship represents either a part-whole relationship or visual evidence supporting the existence of a physical object. We might say that the causality implicit in the conditional independence, i.e., "the whole 'causes' the part" or "the object's interaction with the sensor 'causes' the percept",is an accurate model for vision, in the sense that it is unlikely that the part or percept would occur without the presence of the parent object. The equations for accrual are listed in figure 5.

We have implemented a machine vision Bayes' net only over the portion of the DAG shown in figure 1b. In the following, we explain the representation and algorithms for computation of these prior probabilities. There are two basic kinds of nodes. One type contains competing hypotheses of parts and sub-parts of an object, such as the valves, pipes, connectors, etc., and the other are spatial relationships between these parts, represented in dimensions, ratios of dimensions, and joints between parts. We need to produce priors for nodes of both types. The theory underlying our probabilities is the notion of quasi-invariant observables for physical objects.

## Quasi-invariants and Probabilities

An observable is a measurement repeatable by different observers, i.e. invariant under isometries of the measurement space, or an invariant functional of observables, e.g. the distance between two identifiable points. An observable may be invariant under other transforms, e.g. perspective. There are few perspective invariants that are useful in machine vision, beyond a few which are so fundamental they are overlooked. An observable is quasi-invariant with respect to viewpoint if it is constant to second order under orthographic transforms parameterized by viewing angle on the viewing sphere [Binford 81, 87]. Observables are quasi-invariant under perspective projection if they are constant to second order in the ratio of object dimension to viewing distance on the infinite viewing ball, e.g. if they are invariant under orthographic projection. Observables are quasi-invariant with respect to source orientation or location if they are constant to second order in lighting angle on the unit lighting sphere, i.e. an orthographic transform, or constant to second order in the ratio of object dimension to viewing distance on the infinite lighting ball, perspective lighting trans-forms. Those quasi-invariants discussed thus far have been quasi-invariant under orthographic or perspective projection. These relations are much stronger than stable under general viewpoint, i.e. relations which hold on an open set about a viewpoint. These relations are related to generic invariant, i.e. invariant except on a set of measure zero. We want to generalize that definition from orthographic or perspective projection: an observable is quasi-invariant under a specified set of transformations if it is "approximately invariant over a set with large measure", i.e. over a large fraction of the space of observation. For example, the projected length of a vector is second order in angle about zero; 70% of the angular range has projected length within 30% of true length. We have not formulated "approximately invariant" yet; it relates to a fraction of the meaningful range of variation, a small fraction of one in the length example. It seems straightforward to make definitions for measure for usual cases of parameter spaces. We also call quasi-invariant with respect to object class those observables that are "approximately constant over a set with large measure" with respect to object parameters, e.g. length to width ratio of a GC.

One paradigm for object identification within model-based vision systems is hypothesis generation, verification, and refinement. Hypothesis generation is key, i.e. if the correct hypothesis is included in a small set of hypotheses, there is typically sufficient information to verify the correct hypothesis to great detail. Quasi-invariant metric properties are especially significant for hypothesis generation. We have used quasi-invariants for viewpoint-insensitive prediction in ACRONYM and SUCCESSOR, for generic interpretation of line drawings, and in stereo correspondence. Two quasi-invariants for stereo were analyzed and incorporated into dynamic programming optimization of correspondence measures [Arnold 80].

Values of quasi-invariants under perspective are approximately equal to the values measured from the object in space, hence they are valuable for generating hypotheses of instances of objects. For example, the ratio of lengths of colinear vectors is constant within 10% over 90% of the viewing sphere; it is an orthographic invariant.

Coincidence is a true invariant but it is never observable. Non-coincidence is observable and quasi-invariant under rotation. Non-coincidence is stable for points until they are far enough away from the observer that their apparent distance is below the limiting resolution. If surfaces are non-coincident or non-smooth, these are quasi-invariant observables that are important for segmentation. In [Binford 87] we present a generic observability model for matte reflecting surfaces. Thus, quasi-invariants provide a computational basis for determining prior probabilities for visual events.

Primitive-joints are defined constraints on sets of GCs, and joints in general are defined as sets of primitive-joints. The primary instantaneous (i.e. in a single image) observable about joints are the observed angles between pairs of GCs that are joined at the same primitive joint. Angles between GCs are quasi-invariants. We define prior conditional probabilities for angular relationships between sets of GC's major axes given that they are part of a certain type of primitive joint as follows. Because the probability of observing the angles between each pair of observed GCs is conditionally independent of the probability of observing the angle between any other pair, given the joint all the GCs comprise, we compute the probability of all the angular relationships between the observed-GCs as the product of the conditional probabilities of observed angles between each pair of GCs in the joint. In this paper, we give accrual formulae only for rigid joints. However, by integrating a distribution representing the a priori likelihood of positions over the range of movement for articulated joints, over the probabilities of observability for each intervening position, we can directly extend this theory for articulated joints. This is essentially a convolution, and so issues of computational complexity can be attacked from that basis.

To compute the probability of a observing a given interval of angles between pairs of GCs, we place the pair of GCs in a unit viewing sphere that has its origin at the point of attachment between the GCs. We then determine the area on the surface of the unit viewing sphere from which any of these angles will be observed in an image plane perpendicular to the line of sight determined by a point on the viewing sphere and the origin. Dividing this by $4\pi$, the total area of the unit sphere, we create a probability function on subsets on the unit sphere, and thereby induce a probability function on the space of observed-angles for this pair of GCs. In mathematical notation,

$$p(observing_angle_interval_between_GC_pair|primitive_joint) = \int_{V(a,b)} F(\phi_1, \phi_2) d\phi_1 d\phi_2$$

where $V(a,b) = \{(\phi_1, \phi_2) \; in \; unit \; sphere \; | a \leq observed_angle \leq b\}$ and $F(\phi_1, \phi_2)$ is a distribution over the sphere, expressed in spherical coordinates (elevation, azimuth) representing our prior likelihood of having an observation point at $(\phi_1, \phi_2)$. F is used to represent, for example, that an industrial robot is likely to be looking at a workbench from a certain interval of depression angles depending on the height and distance of the sensor from the workbench.

Figure 6 pictures this situation for a pair of GCs and shows the value of the cosine of the observed angle for a pair of GCs with an angle $\theta$ between them in standard position, observed from a point $(\phi_1, \phi_2)$ on the unit viewing sphere. We have provided numerical solutions for small intervals of angles. Figure 7 shows conditional prior probability of observed angle given that the true angle is 90 degrees. Because each location on the sphere corresponds to a unique observed angle for a fixed pair of GCs, we can compute the area on the sphere for observing an interval of angles by summing the areas corresponding to the appropriate small intervals.

Recall that the other sort of prior probability we had to determine was the probability of a joint or part given a parent part or parent joint. We propose to compute these by counting the number of parts or joints for which the child node is a constituent, and multiplying the fraction of times the child occurs times the number of children of that type that occurs in the parent part. This uses our knowledge of the domain directly. For a given manufacturing job, for instance, this amounts to having a priori knowledge of the potentially required parts. For each alternative hypothesis in a child node, we have a priori knowledge of the set of parts (joints) of which it is a sub-joint (sub-part). Take the union of all the parent parts obtained for the set of hypotheses in the child node. Then determine the a priori probability of each child hypothesis given a $parent_i$ as

$$\frac{K[\chi(child_hyp(parent_i)) * number_child_hyps_in(parent_i)]}{\sum_{j \neq i}[\chi(child_hyp(parent_j)) * number_child_hyps_in(parent_j)]}$$

where $\chi(child_hyp)$ is a function that is 1 if the child occurs in the parent, and 0 otherwise, and $K$ is a factor normalizing over all child hypotheses. Note that one of the child hypotheses must be "other" in order to represent ignorance in a probabilistic fashion. The un-normalized probability of other given a $parent_j$ is

$$\frac{\sum_i [1 - \chi(child_hyp_i(parent_j))]}{number_child_hypotheses}$$

## Results

Figures 8a and 8b show photographs of two objects, a valve and an elbow-joint. Figure 8c shows the results of applying an edge operator to the valve image. Figure 8d shows the results of manually associating observed edges into ribbons that were classified as straight and octagonal, respectively. The inference of the ribbons triggers the generation of the hypotheses of joints. The DAG for the joint consisting of the pair of rectangular GCs is pictured in figure 9a. After posting several joint hypotheses, we can infer the possible presence of the valve, elbow, or other object. The full DAG is shown in figure 9b. The values of probabilities propagated up the DAG are listed in the nodes.

Figure 10 lists the values for top level object hypotheses accrued on the same observed evidence as the total set of possible objects varies. As expected, when the elbow hypothesis is not generated, the probability of the valve increases. When the true hypothesis is omitted, the probability of the "other" hypothesis goes up correspondingly.

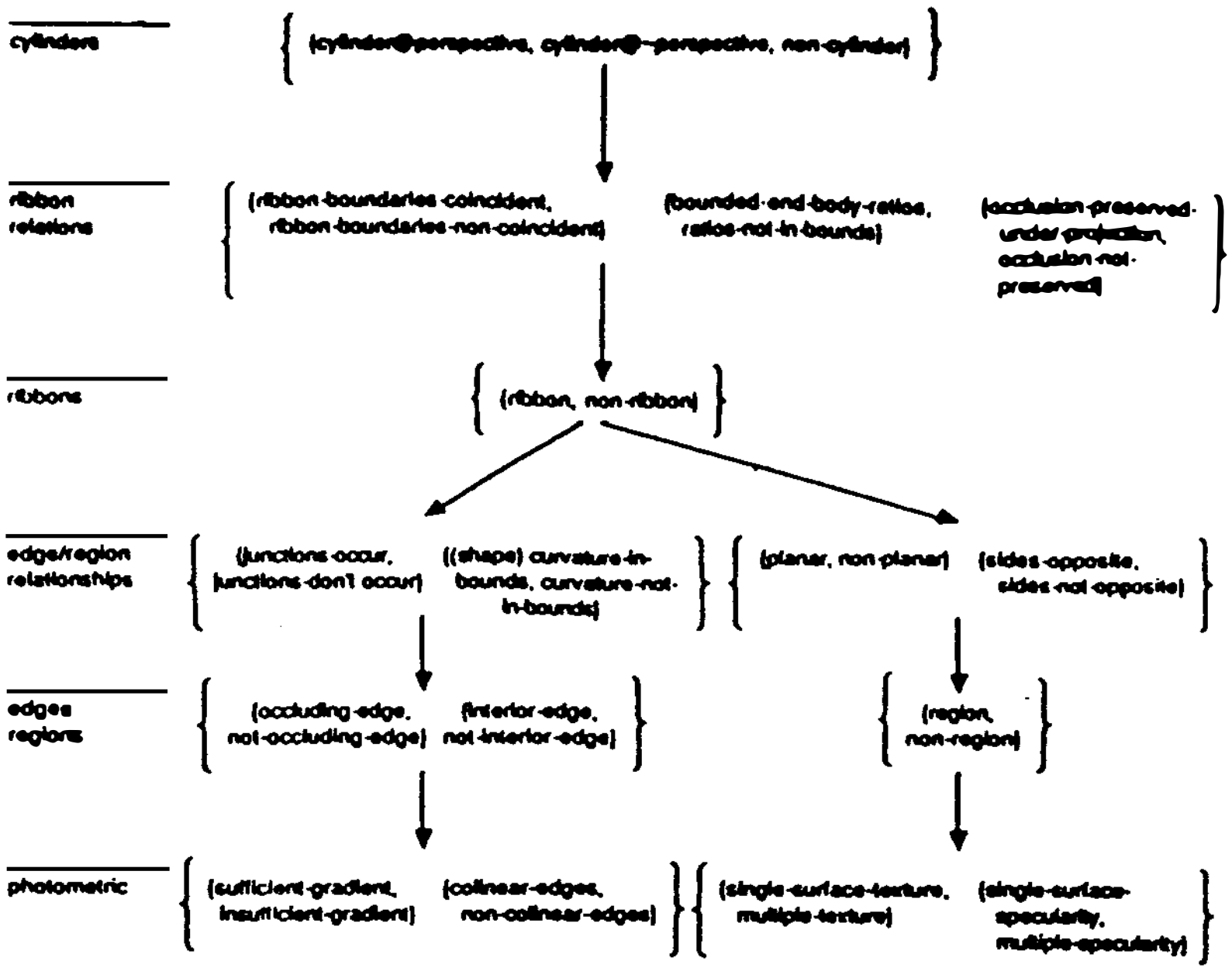

Figure 1a: DAG up to level of GCs.

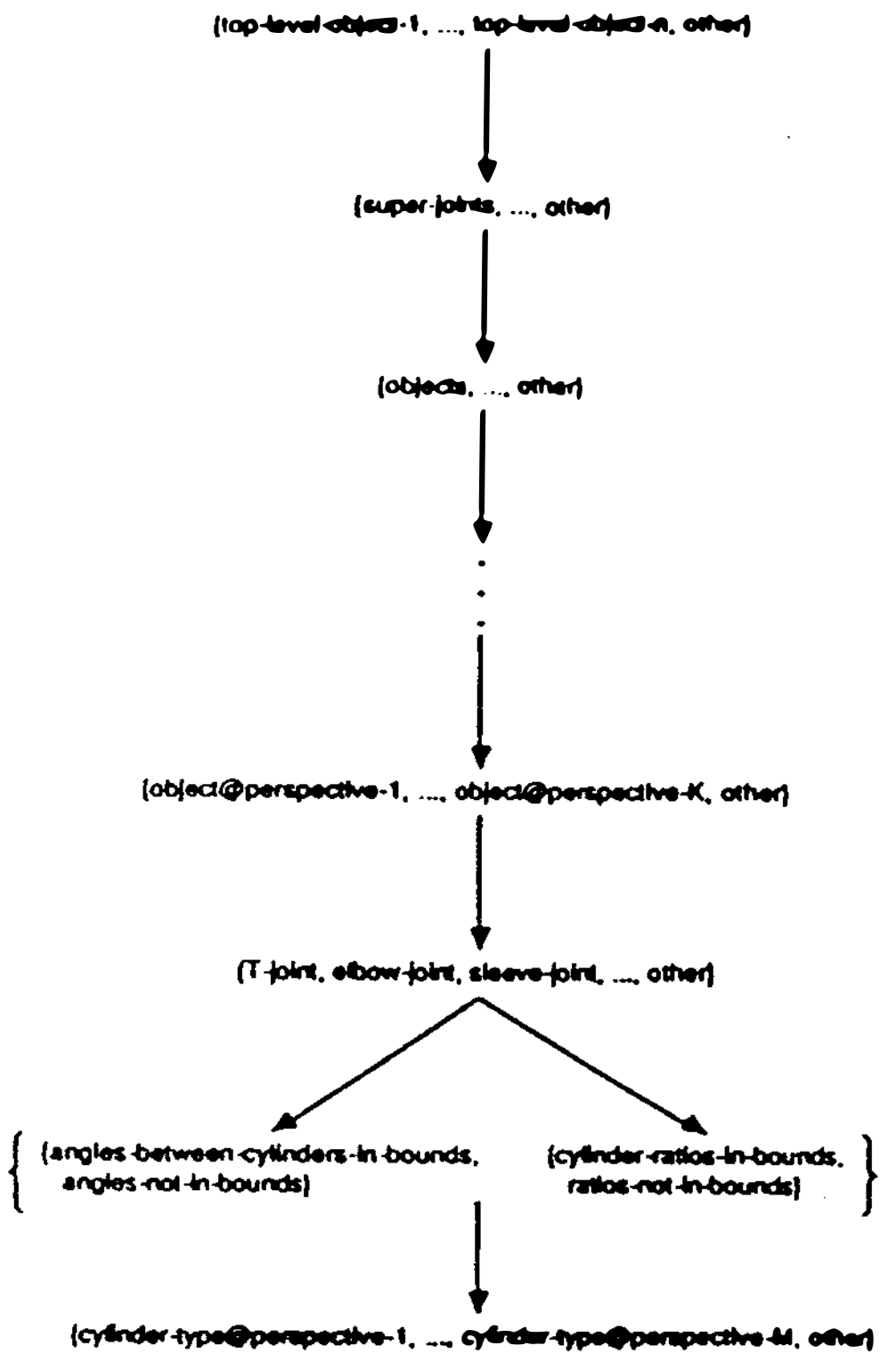

1b: object-level DAG

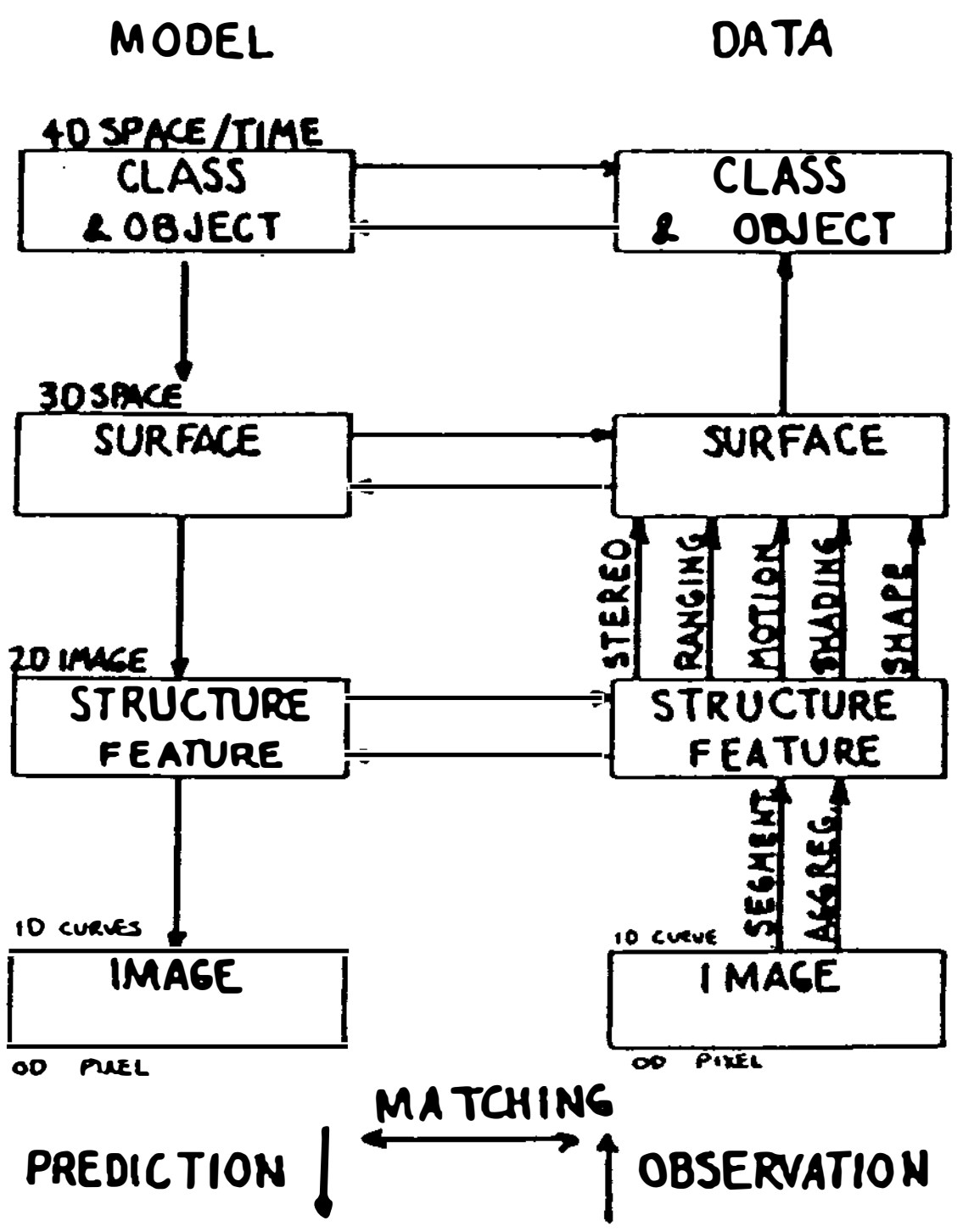

Figure 2: Machine Vision Inference Structure

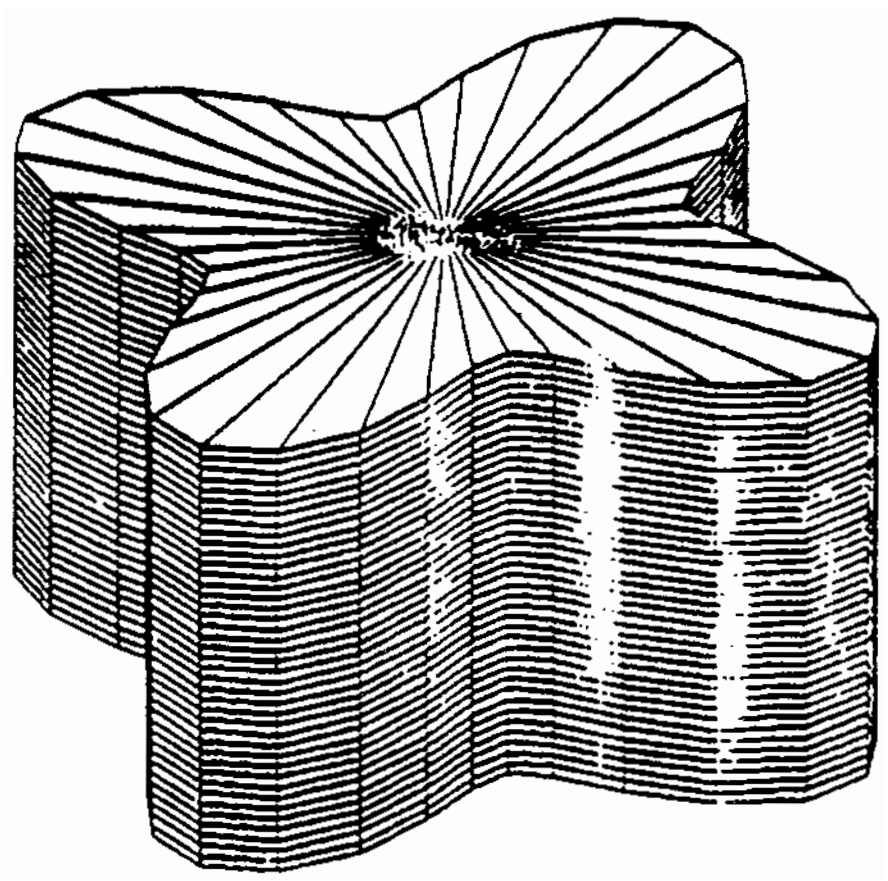

Figure 3a: True cylinder
Sweeping rule is constant.

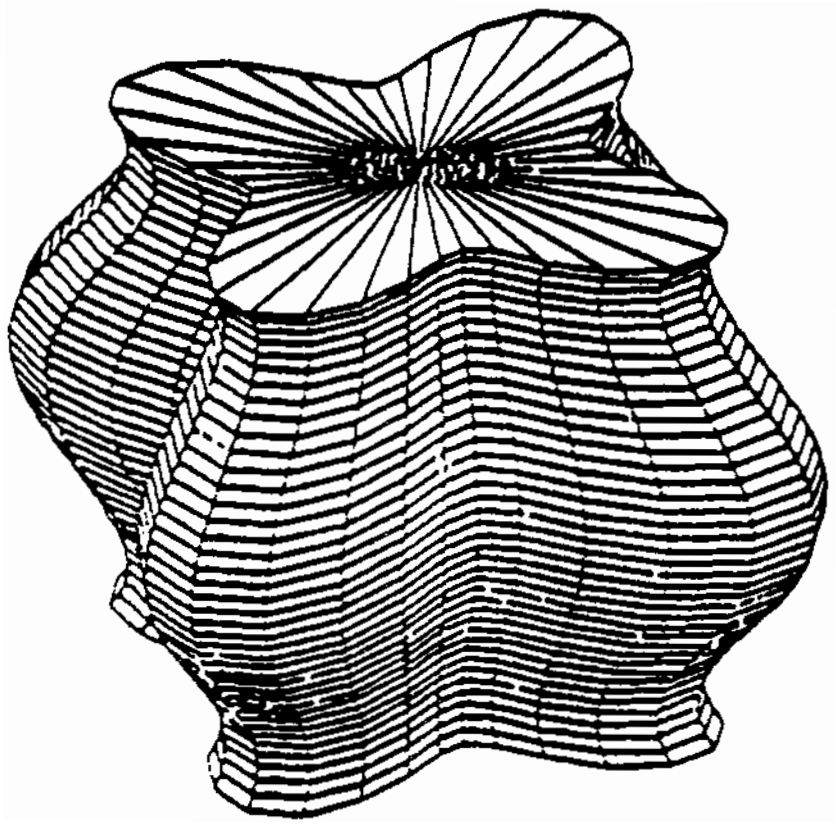

3b: generalized cylinder

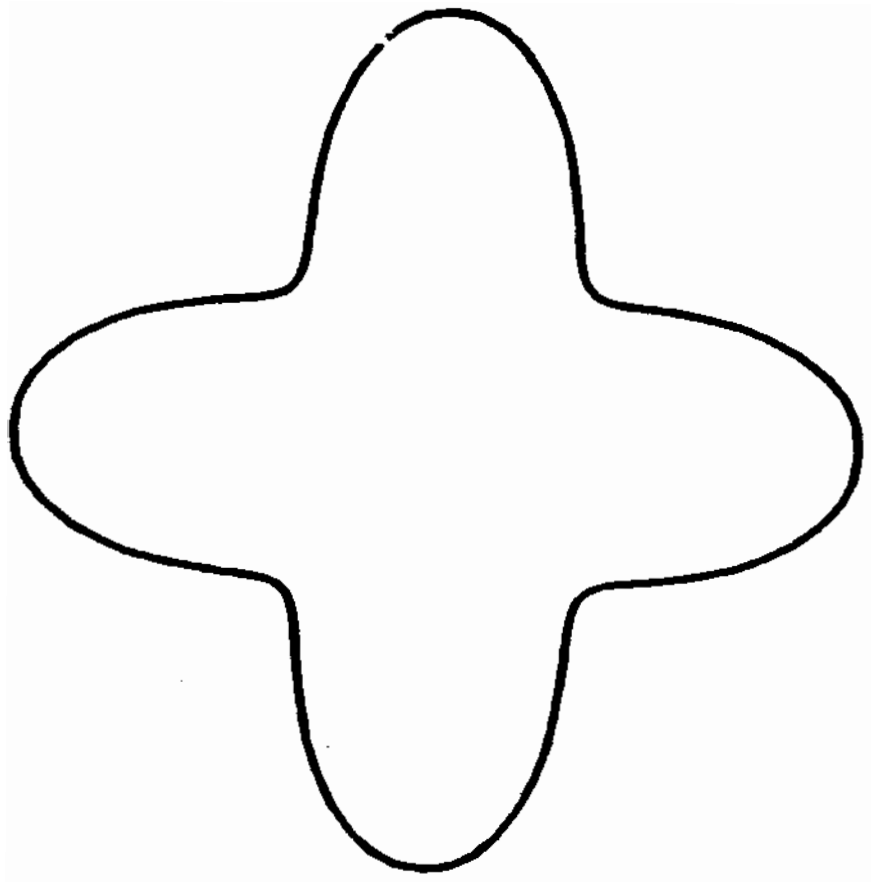

3c: cross section for 3a, 3b

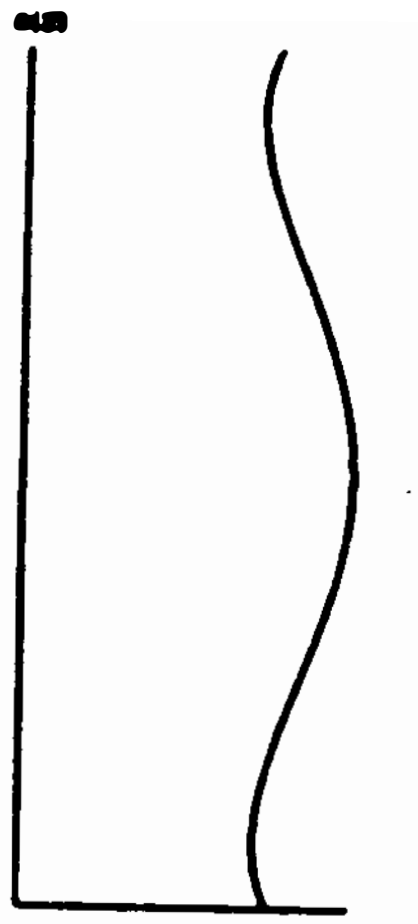

3d: sweeping rule for 3b

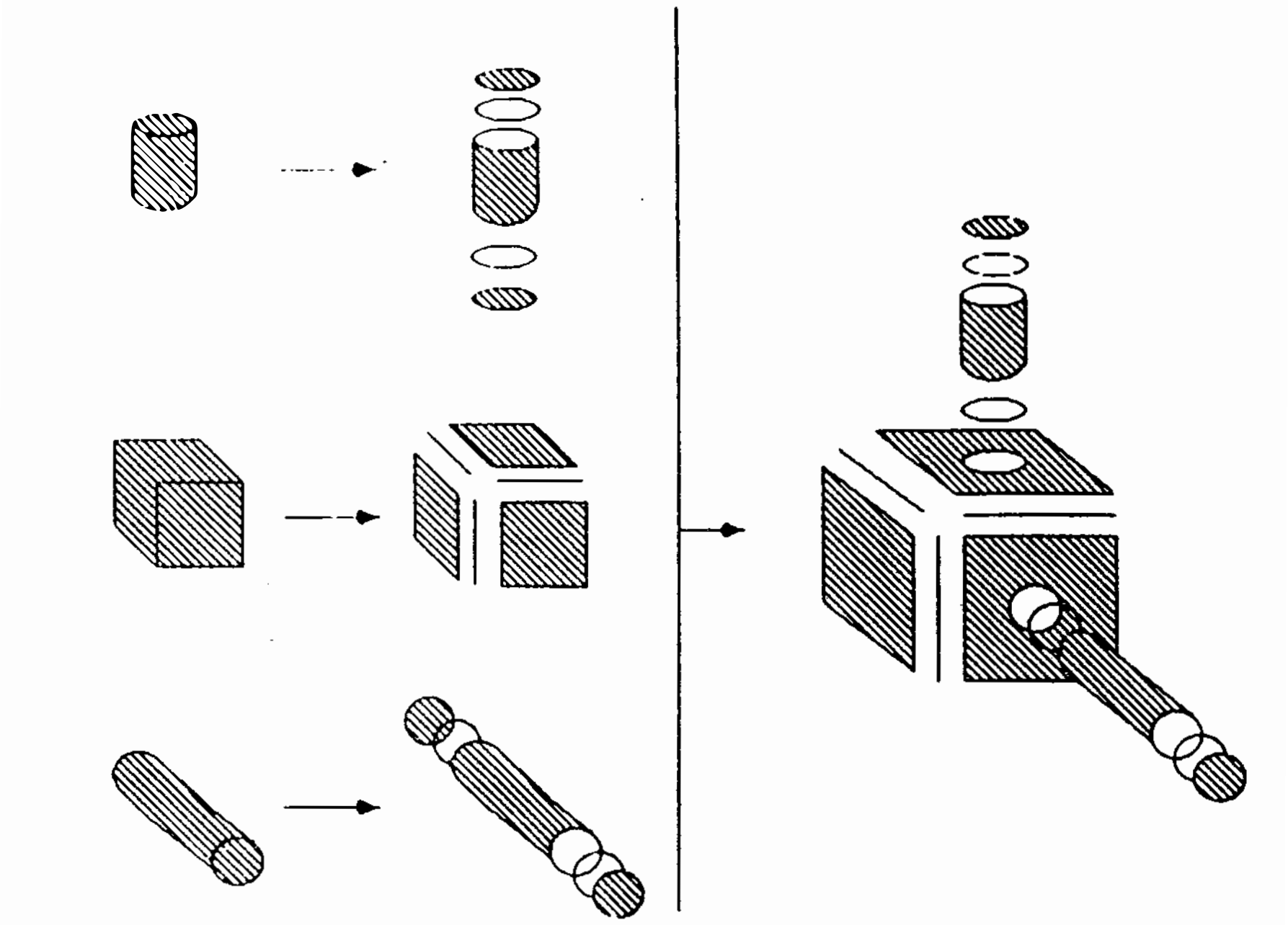

4a. object parts 4b. surfaces and edges 4c. surfaces and edges with holes and hidden faces.

Figure 4: Prediction

*Given DAG:*

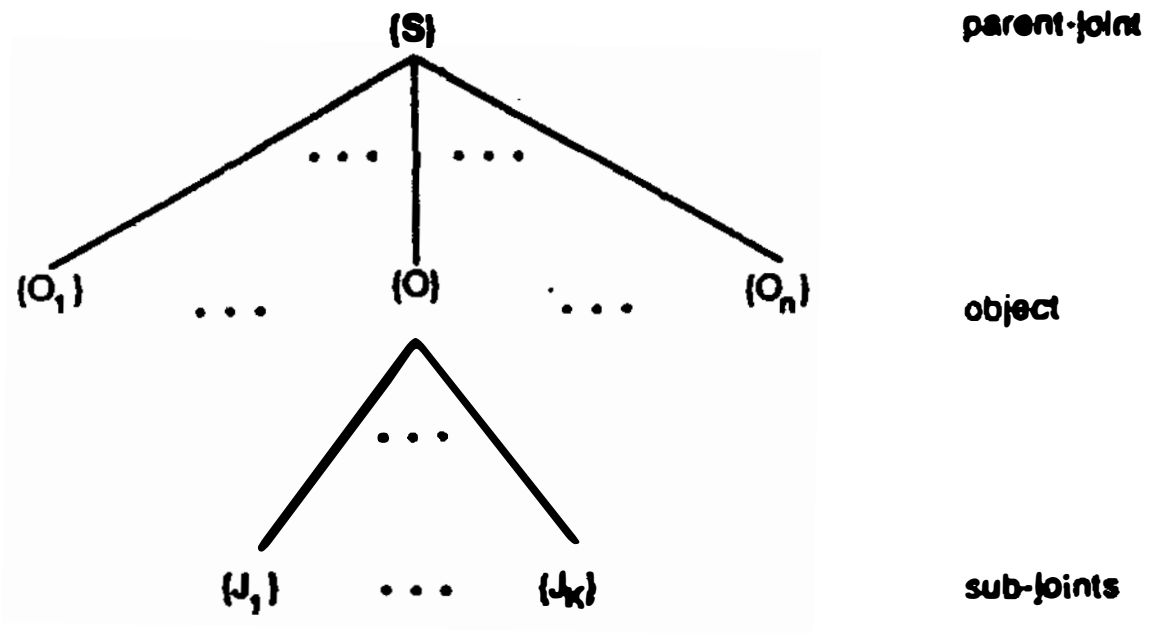

**When the node $\{O\}$ is activated:**

1) Fetch $m(S)$ and $\{m(J_i)\}$, the current messages from parent and children

2) Compute $C_r = \prod_i m(J_i)_r$ for each hypothesis $O_r$ of $O$ (i.e., each $m(J_i)$ is a vector $\{m(J_i)_1, \ldots, m(J_i)_l\}$ where each $m(J_i)_r$ is the probability contributed to the r-th hypothesis of $\{O\} = \{O_1, \ldots, O_r, \ldots\}$)

3) Compute $P_r = \beta \sum_j p(O_r \mid S_j)\, m(S_j)$ where $\beta$ is normalizing over $\{O\}$ and $p(O_r \mid S_j)$ is the prior probability value

4) Current belief in $O_r$ is $\alpha C_r P_r$ where $\alpha$ is normalizing over $\{O\}$

5) Send $S_j$ the message $\sum_i p(O_i \mid S_j) C_i$

6) Send $J_K$ the message $\dfrac{\alpha C_r P_r}{m(J_K)_r}$

**Figure 5: Bayesian Probability Propagation**

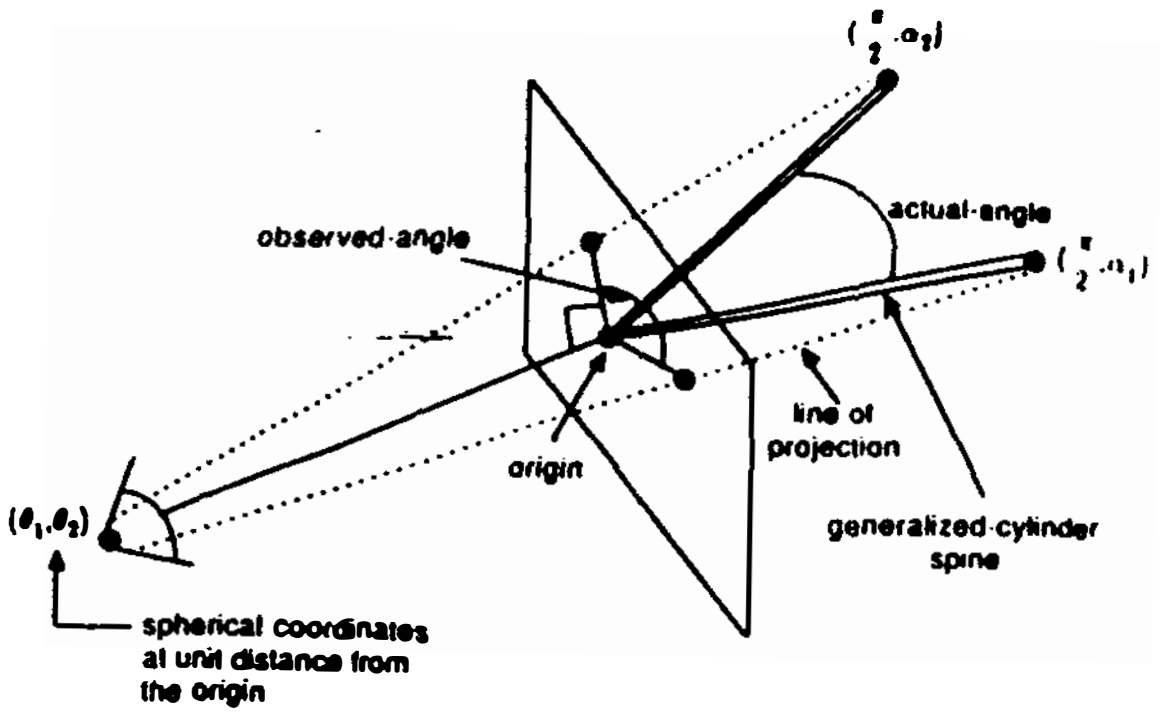

$$\text{actual-angle} = \theta_2 - \theta_1$$
$$A = \cos\theta_1$$
$$B = \sin\theta_1$$
$$C = \cos\theta_2$$
$$D = \sin\theta_2$$

$$\text{observed-angle} = \text{arc-cosine}\Big[\,\big|(A(1-\sin^2\phi_1\cos^2\phi_2) + \tfrac{B}{2}\sin^2\phi_1\sin^2\phi_2)\cdot(C(1-\sin^2\phi_1\cos^2\phi_2) + \tfrac{D}{2}\sin^2\phi_1\sin^2\phi_2) + (-\tfrac{A}{2}\sin^2\phi_1\sin^2\phi_2 + B\sin^2\phi_1\sin^2\phi_2 + B)\cdot(-\tfrac{C}{2}\sin^2\phi_1\sin^2\phi_2 + D\sin^2\phi_1\sin^2\phi_2 + D) + (1-\sin^2\phi_1\cos^2\phi_2)(B\sin\phi_1\sin\phi_2 - A\sin\phi_1\cos\phi_2)\cdot(D\sin\phi_1\sin\phi_2 - C\sin\phi_1\cos\phi_2)\big| \;/\; \big|(1+2\sin^2\phi_1(B^2\sin^2\phi_2 - A^2\cos^2\phi_2) + \sin^2\phi_1(B\sin\phi_2 - A\cos\phi_2)^2)\cdot(1+2\sin^2\phi_1(D^2\sin^2\phi_2 - C^2\cos^2\phi_2) + \sin^2\phi_1(D\sin\phi_2 - C\cos\phi_2)^2)\big|^{1/2}\,\Big]$$

**Figure 6: Observed Angle Computation**

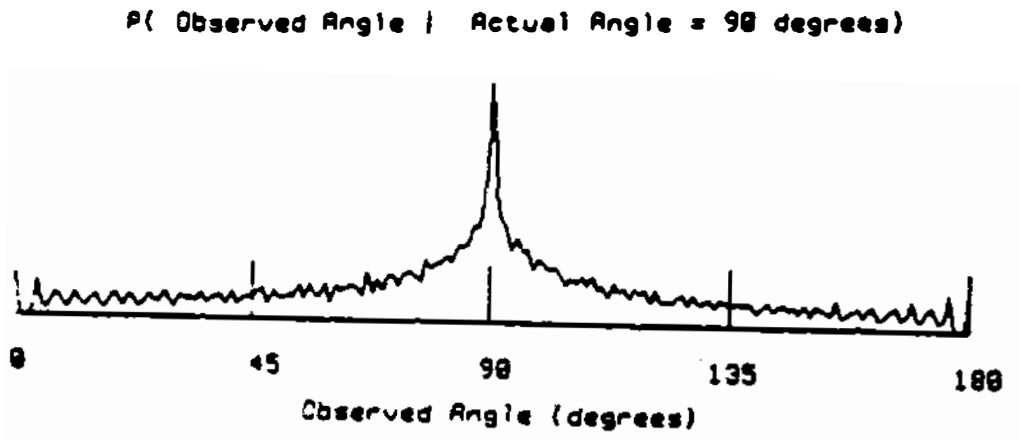

← **Figure 7: Quasi-invariant probability of observed angles given 90 degree actual.**

**Figure 9a. Instantiated DAG for Rectangular Prism Joint** →

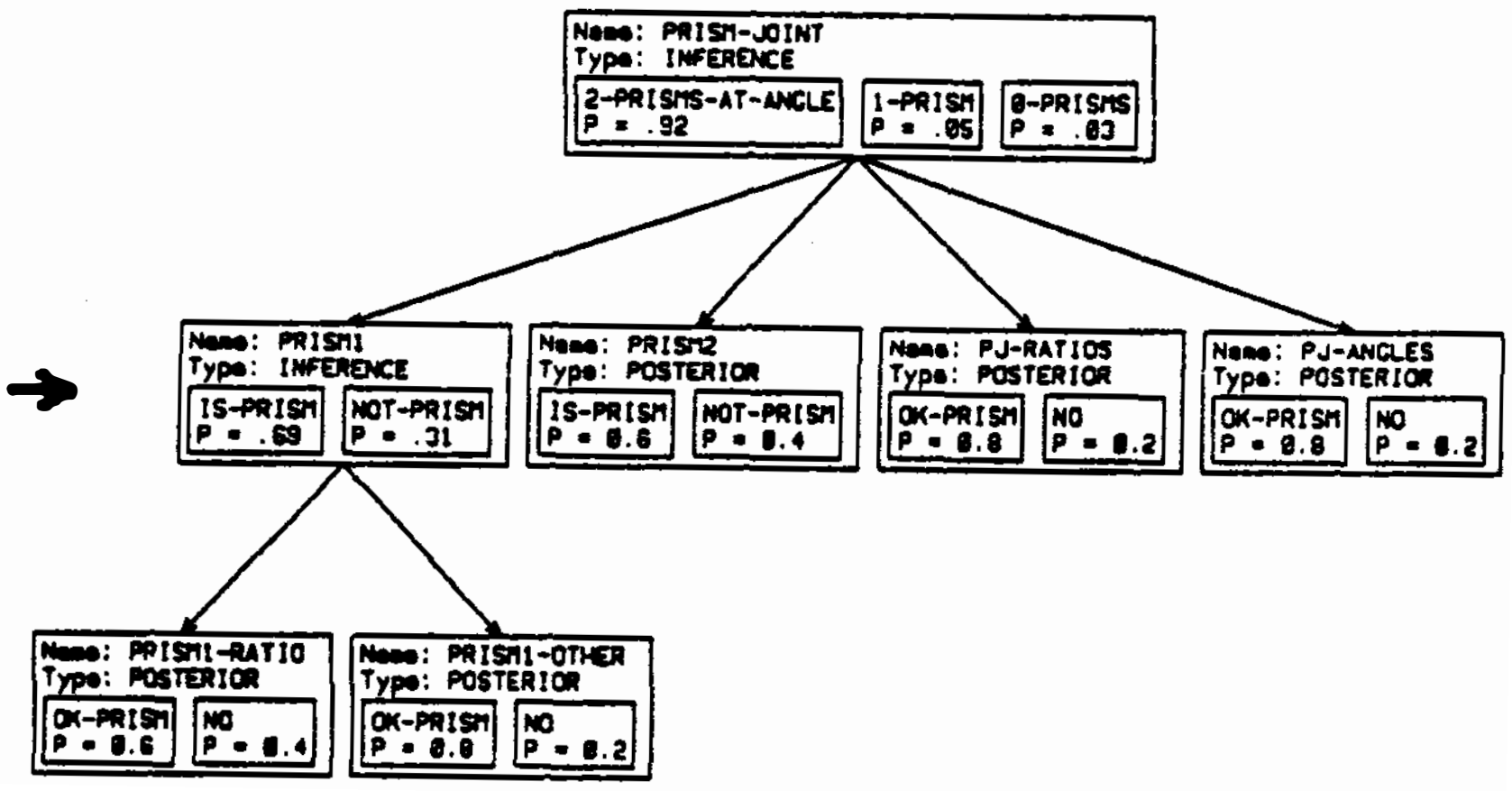

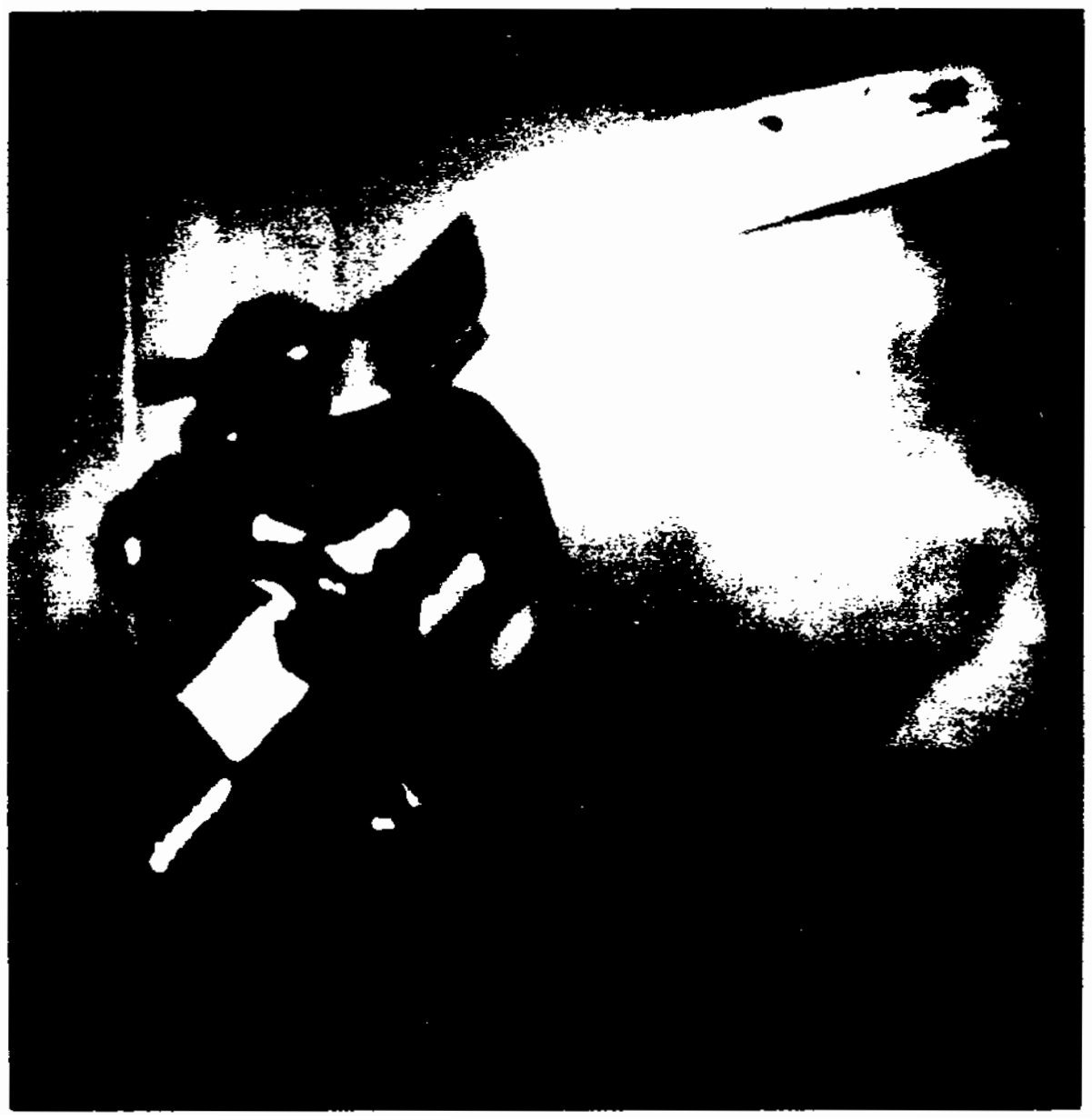

Figure 8a. Valve image

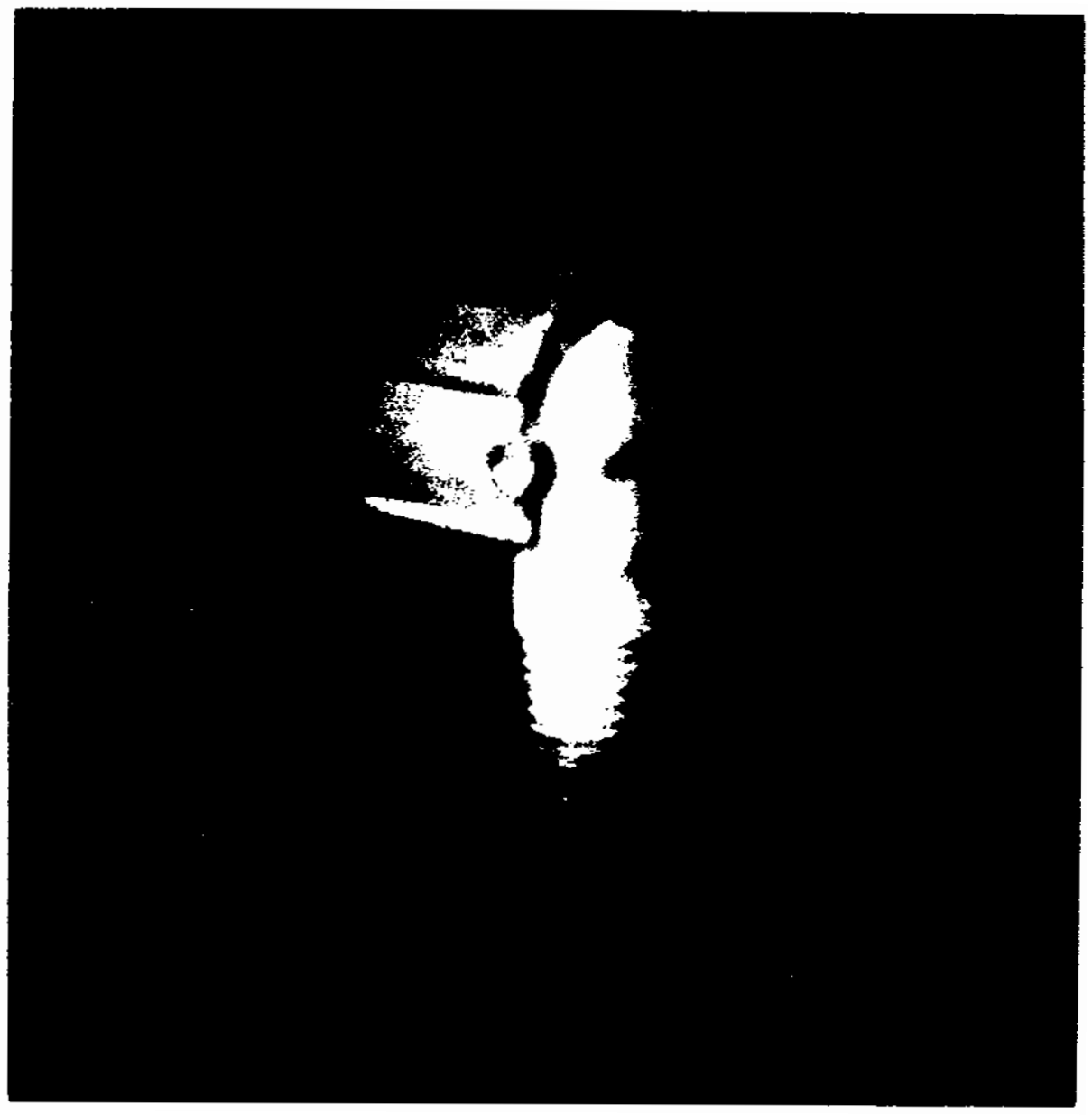

8b. Elbow Joint Image

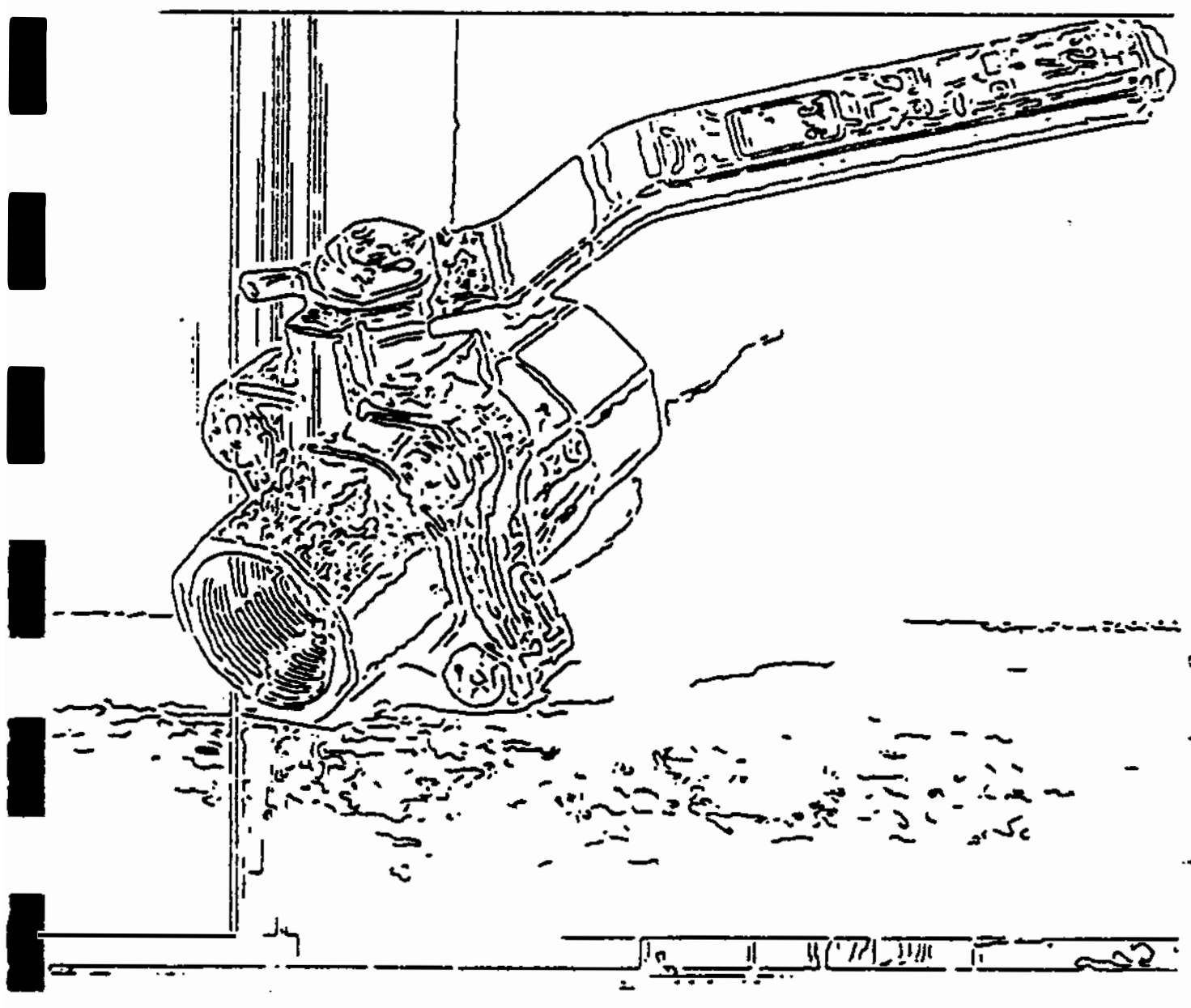

8c. Edges from Valve Image

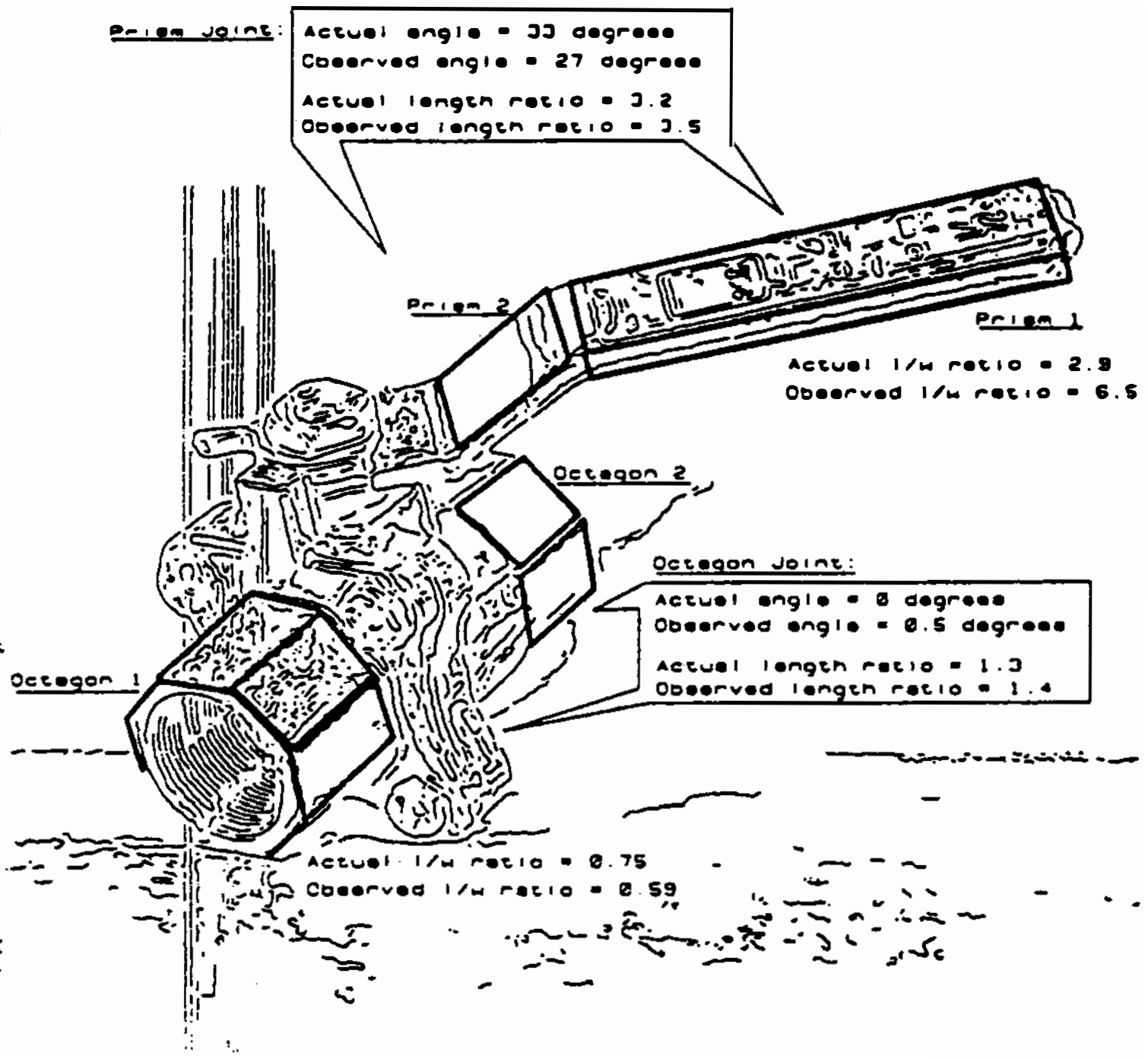

8d. Ribbon inference and quasi-invariant measures

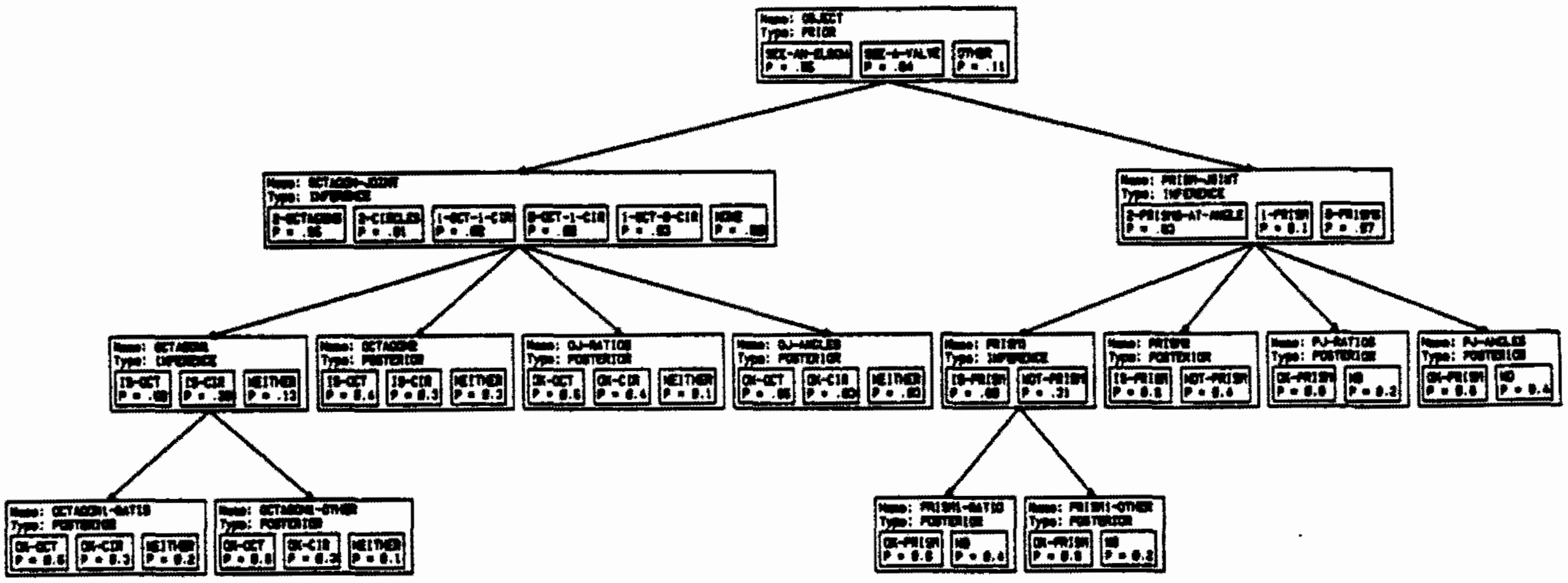

**Figure 9b. Instantiated DAG for Object Hypotheses.**

| | Hypotheses | | |
|---|---|---|---|
| | Elbow | Valve | Other |
| Probability | .06 | .83 | .11 |
| | .37 | --- | .63 |
| | --- | .88 | .12 |
| | .08 | .92 | --- |

**Figure 10. Evidence for top level hypothesis sets.**